\documentclass[conference,compsoc]{IEEEtran}
\usepackage{graphicx}
\usepackage{float}

\ifCLASSINFOpdf
\else
\fi
\begin{document}
	
	\title{SportsTrack: An Innovative Method for Tracking Athletes in Sports Scenes}
	\renewcommand{\thefootnote}{\fnsymbol{footnote}}
	
	\author{
		\IEEEauthorblockN{Jie Wang\IEEEauthorrefmark{1}, Yuzhou Peng\IEEEauthorrefmark{2}, Xiaodong Yang, Ting Wang, Yanming Zhang}
		\IEEEauthorblockA{ \\
			\IEEEauthorrefmark{1} First author\\
			\IEEEauthorrefmark{2} Corresponding author
		}
	}
	
	
	% compsoc conferences compared to traditional conferences), use this
			%Telephone: (800) 555--1212, Fax: (888) 555--1212}


	\maketitle
	
	\begin{abstract}
		The SportsMOT dataset aims to solve multiple object tracking of athletes in different sports scenes such as basketball or soccer. The task is challenging because of the unstable camera view, athletes' complex trajectory, and complicated background. Previous MOT methods \cite{DBLP:journals/corr/abs-2010-12138,DBLP:journals/corr/abs-2003-13870,DBLP:journals/corr/abs-2006-06664,DBLP:journals/corr/abs-2012-15460,DBLP:journals/corr/abs-2103-08808,DBLP:journals/corr/abs-2004-01888} can not match enough high-quality tracks of athletes. To pursue higher performance of MOT in sports scenes, we introduce an innovative tracker named SportsTrack, we utilize tracking by detection as our detection paradigm. Then we will introduce a three-stage matching process to solve the motion blur and body overlapping in sports scenes. Meanwhile, we present another innovation point: one-to-many correspondence between detection bboxes and crowded tracks to handle the overlap of athletes' bodies during sports competitions. Compared to other trackers such as BOT-SORT \cite{aharon2022bot} and ByteTrack \cite{DBLP:journals/corr/abs-2110-06864}, We carefully restored edge-lost tracks that were ignored by other trackers. Finally, we reached state-of-the-art tracking score (76.264 HOTA) in the SportsMOT dataset.
	\end{abstract}
	
	%%%%%%%%% BODY TEXT
	\section{Introduction}
	\label{sec:intro}

\begin{figure}
		\centering
		\includegraphics[width=0.5\textwidth]{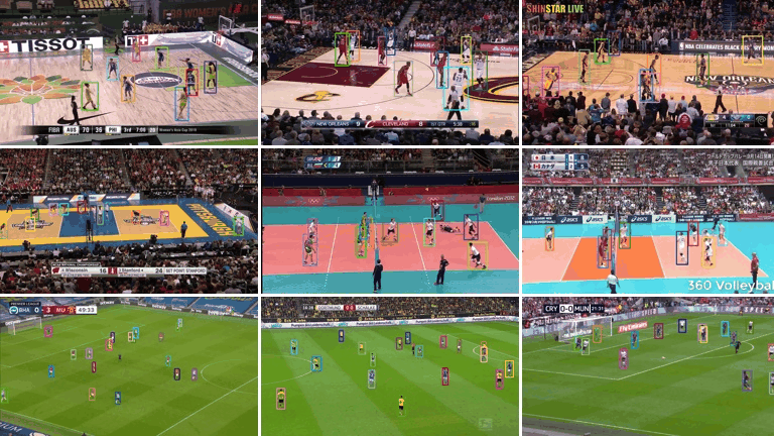}
		\caption{Overview of different sports scenes.}
		\label{fig1}
	\end{figure}

\begin{figure}
		\centering
		\includegraphics[width=0.5\textwidth]{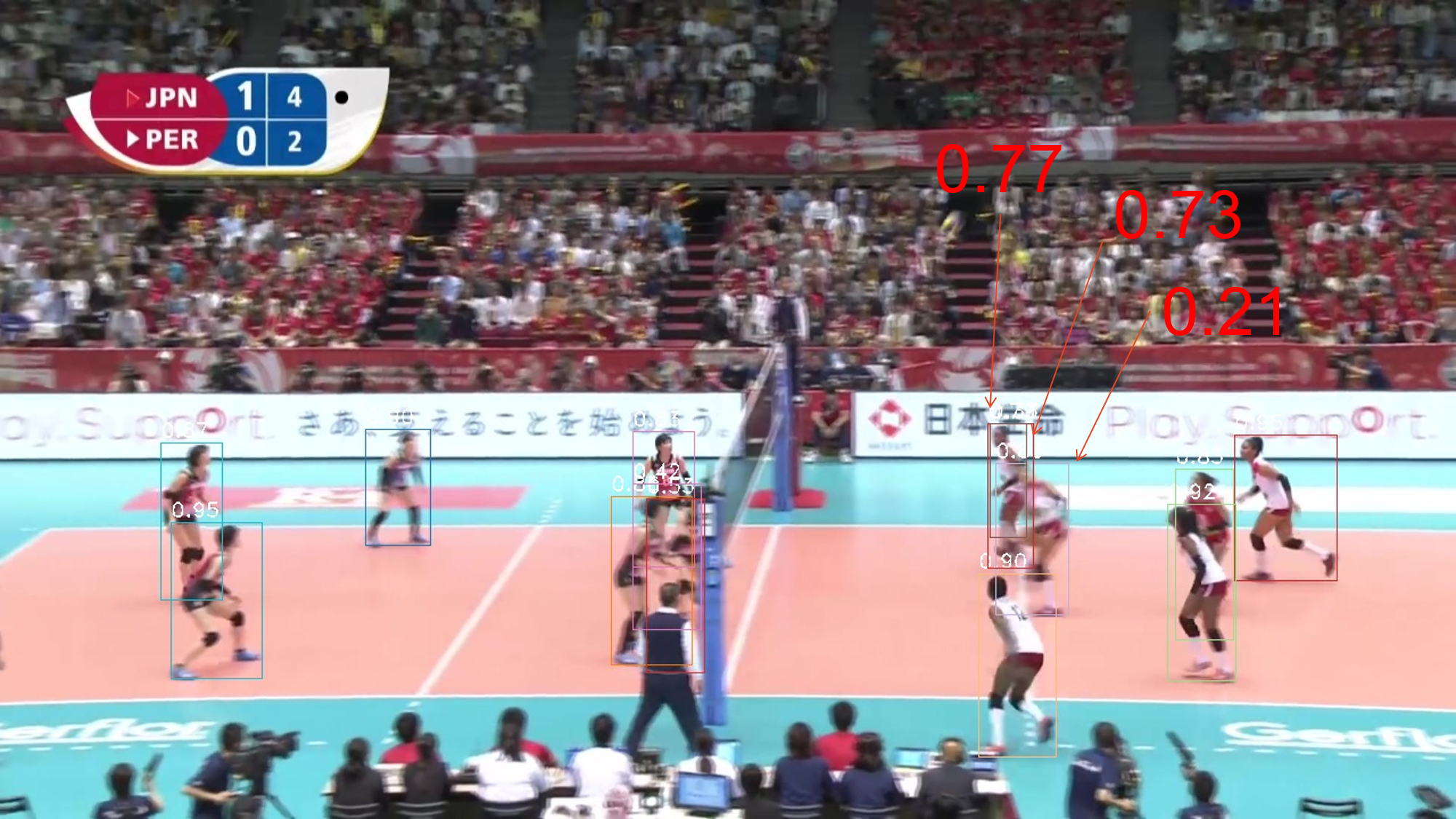}
		\caption{Motion blur of volleyball athletes}
		\label{fig2}
	\end{figure}

	The MOT (Multiple Object Tracking) is one of the fascinating directions of computer vision. The workflow of MOT is usually composed of three steps: (1) Detection (2) Similarity calculation (3) Matching. In the entire MOT process, retrieving high-quality detection and resolving missing detection is crucial for tracking, and the overlap of targets can also affect the tracking results. How to distinguish targets and track them thus becoming challenging. In different sports scenes, detecting athletes from viewers and judgments and tracks matching can be extremely difficult since the rapid motion of athletes and complex environments like Figure \ref{fig1}. We can see from Figure \ref{fig2}, the right-side athletes corresponds to multiple detection bboxs with confidence value 0.77 and 0.73, and the purple bbox with confidence value 0.21 has single blurred athletes. If we utilize the matching stratagy similar to Bytetrack, the athlete's track will match high-confidence bbox first and then match low-confidence bbox, for athletes in the purple detection box, because the confidence of the purple detection box is low, the athletes will match the 0.77 confidence red detection box with deviation instead of accurately detecting the purple detection box of the athletes, thus becoming a wrong tracking sequence. So we need to handle this situation carefully. 
	
\begin{figure}
		\centering
		\includegraphics[width=0.4\textwidth]{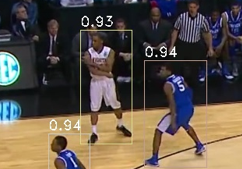}
		\caption{Detection results without body overlapping}
		\label{fig3}
	\end{figure}

\begin{figure}
		\centering
		\includegraphics[width=0.4\textwidth]{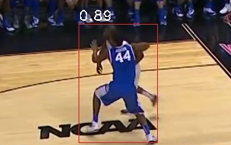}
		\caption{Detection result with body overlapping}
		\label{fig4}
	\end{figure}

	Meanwhile, since those sports scenes are fierce, athletes can get very close during physical confrontations and cause fewer detection results for the overlapping of athletes' bodies. like Figure \ref{fig3} and Figure \ref{fig4}, Figure \ref{fig3} illustrates two separate athletes and each detection bbox corresponds to one athlete. In Figure \ref{fig4}, athletes with a blue shirt and white shirt are moving together and the blue shirt athlete is keeping off the white shirt athletes and detection algorithm can not detect both of them, and this will interrupt the white athlete's tracking process and cause lower tracking score.

\begin{figure}
		\centering
		\includegraphics[width=0.4\textwidth]{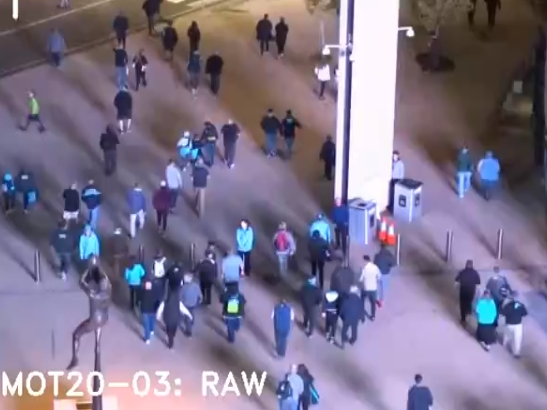}
		\caption{Pedestrian in MOT20 dataset}
		\label{figpede}
	\end{figure}

\begin{figure*}
		\centering
		\begin{minipage}[t]{0.3\linewidth}
			\centering
			\includegraphics[scale=0.3]{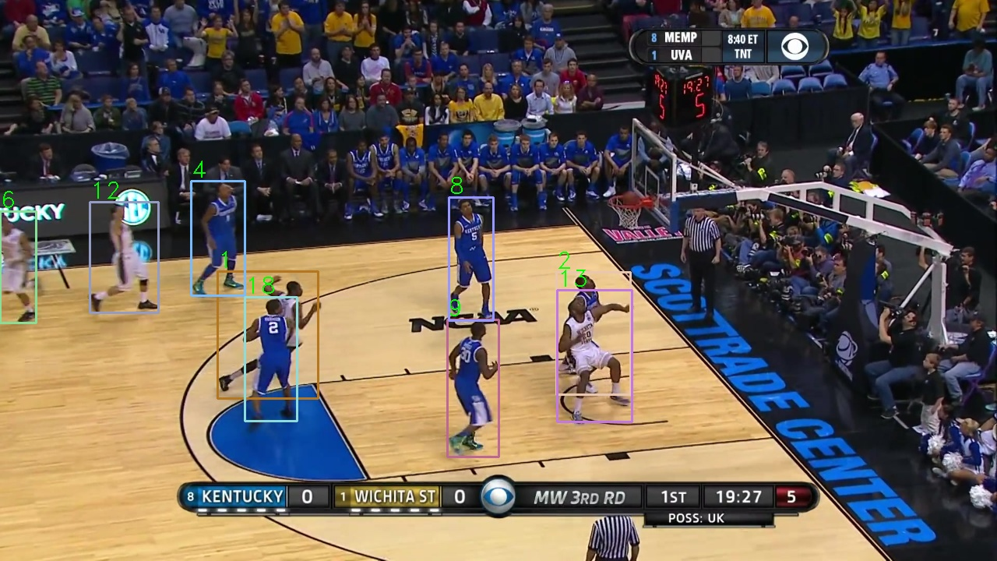}
			\caption{White-shirt athlete with id 6 frame-out from left side.}
			\label{fig5.1}
		\end{minipage}%
		\begin{minipage}[t]{0.3\linewidth}
			\centering
			\includegraphics[scale=0.3]{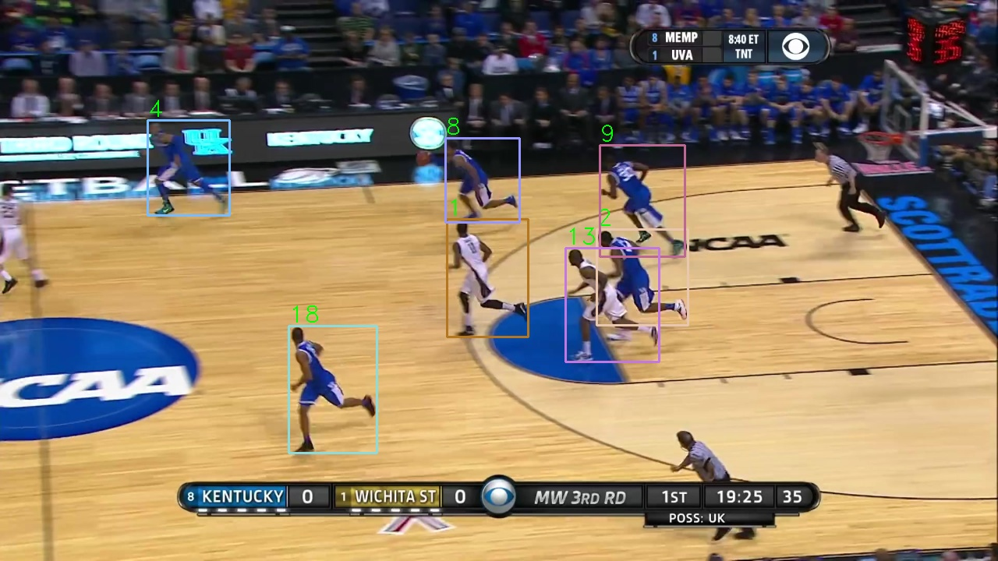}
			\caption{White-shirt athlete with id 6 frame-in from left.}
			\label{fig5.2}
		\end{minipage}%
		\begin{minipage}[t]{0.3\linewidth}
			\centering
			\includegraphics[scale=0.3]{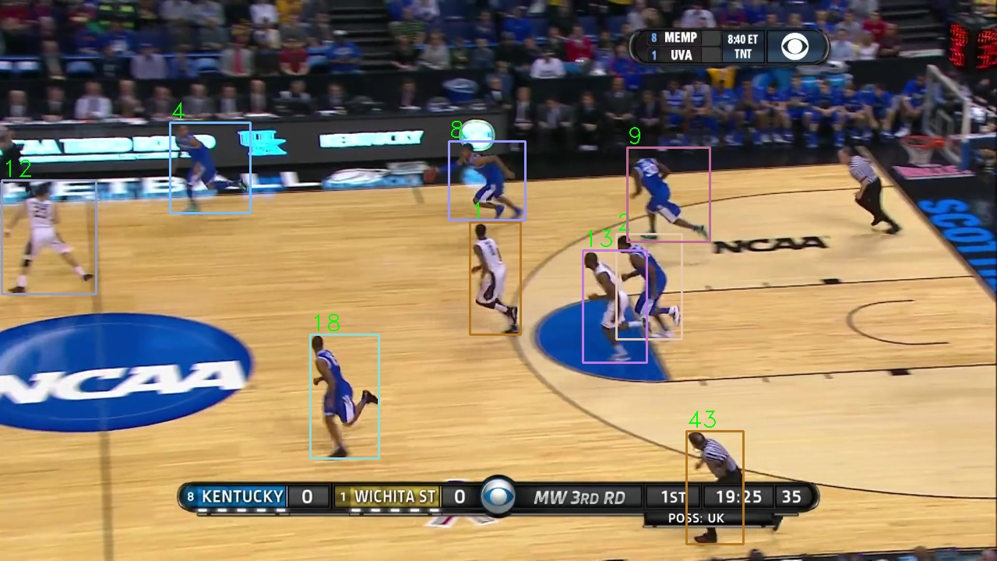}
			\caption{White-shirt Athlete with id 6 switch to id 12.}
			\label{fig5.3}
		\end{minipage}
		
	\end{figure*}

	The dataset of sports scenes is quite different from ordinary MOT datasets details will be illustrated in Figure \ref{fig5.1} \ref{fig5.2} \ref{fig5.3}, Figure \ref{figpede} is picture from MOT20 pedestrian dataset \cite{DBLP:journals/corr/abs-2003-09003}, once the pedestrians are moving out the view of the camera, most of them will not appear again in the frame, it's highly impossible for them to come back to the camera view. However, for sports scenes, athletes can appear in the frame multiple times, how to distinguish and restore tracklets is also crucial for sports MOT
	results.
	During the analysis of sports scene's data, we found that it is necessary for us to match as many high-quality targets as possible. First. Since matching bboxes and tracklets with high and low confidence values (similar to byte track) is not enough for filtering. In sports scenes like Figure \ref{fig2}, low confidence bboxes do not mean low quality (since the motion blur of athletes) and high confidence detection results do not mean high quality (since one athlete corresponds to many bboxes). We will match our detection bboxes and tracklets with an optimized tracker compared with some pedestrian trackers like Bytetrack \cite{DBLP:journals/corr/abs-2110-06864}, the first stage will utilize ReID and IOU distance as standard to calculate the similarity between bboxes and tracks with a stricter threshold, the second stage will use high confidence bboxes to matching tracklets, by calculating ReID and IOU distance,  The last stage, low confidence bboxes will be matched by calculating IOU distance only. The first stage will maximize the probability of retrieving as many high-quality detection bboxes as possible. The second and third stages are similar to strategy of Bytetrack \cite{DBLP:journals/corr/abs-2110-06864}, we will use high and low-confidence bboxes to match tracklets, low confidence bboxes will have more probability to match tracklets. In this way, we will generate high-quality tracklets in the matching process.
	
	For athletes with overlap, it is strenuous for detection algorithms to detect all targets. Lost of detection results will affect the tracking process. The ordinary strategy of matching is one detection bbox can correspond to one tracklets. In this paper, we will introduce the "crowded track", and "crowded track" will be overlapped tracks, and for crowded track and detection bboxes, we allow one-to-one correspondence between them, in this way we will not miss tracks and keep the tracking process consecutive.
	
\begin{figure*}
		\centering
		\includegraphics[width=0.8\textwidth]{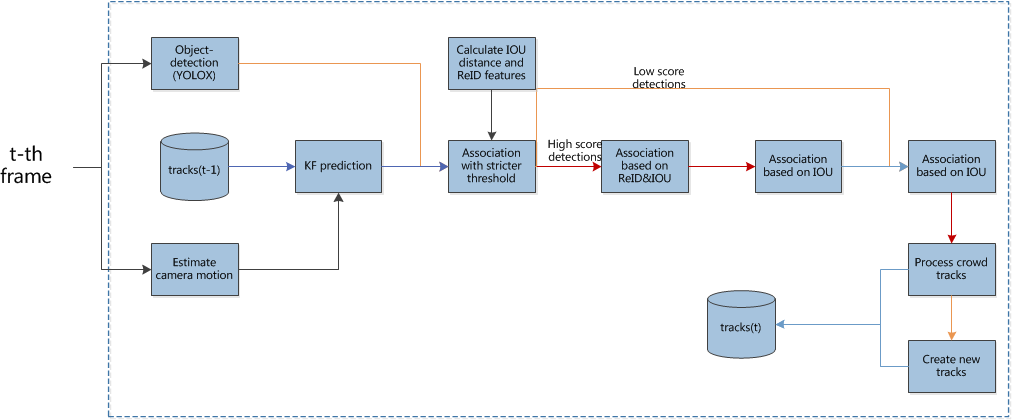}
		\caption{Workflow of SportsTrack}
		\label{fig_work}
	\end{figure*}

	To pursue higher performance of MOT in sports scenes, we introduce an innovative tracker named SportsTrack, we utilized YOLOX\cite{yolox2021}, a strong detection algorithm to detect targets. Then detection bboxes will be used to generate "crowded tracks", and we propose Sports Matching, a three-stage matching strategy. For those bboxes that do not match any tracks, we will generate new tracks with a strict standard. The workflow of our tracker has illustrated in Figure \ref{fig_work} Finally, we reached 76.264 HOTA in the SportsMOT test dataset. We hope SportsTrack will be used in practical applications in the future.
	
	\section{Related Work}
	\subsection{Detection}
	The detection algorithm is an essential part of MOT. The performance of object detection evolving rapidly. The famous 
	two-stage detection algorithm RCNN and its divergence of it \cite{DBLP:journals/corr/GirshickDDM13,DBLP:journals/corr/Girshick15,DBLP:journals/corr/RenHG015} have greatly improved the performance of the detection algorithm. The popular detection algorithm YOLO and its derivatives \cite{DBLP:journals/corr/abs-1804-02767,DBLP:journals/corr/abs-2004-10934} can produce high-quality detection results while having high efficiency, and they have been used in a amount of trackers \cite{DBLP:journals/corr/abs-2104-00194,DBLP:journals/corr/abs-2010-12138,DBLP:journals/corr/abs-2104-09441,DBLP:journals/corr/RenHG015} for detection tasks.
	
	Since the targets in sports scenes have motion blur and result in the missing of detection targets. Base on previous methods \cite{DBLP:journals/corr/abs-1811-05340,DBLP:journals/corr/abs-1801-09823}, we will maximize the use of previous frame's information to optimize the performance of detection.
	
	Detection-by-tracking is also vital for retrieving more accurate results of detections, some trackers \cite{DBLP:journals/corr/abs-2104-09441,DBLP:journals/corr/abs-1811-11167} use tracked bboxes in previous frames for improving the following frame's feature representation. Other trackers utilize Kalman filter \cite{kalman1960} or single object tracking \cite{DBLP:journals/corr/BertinettoVHVT16} for the location prediction of tracks in following frames and merge predicted bboxes with detection bboxes to strengthen the detection results. In our tracker, we will improve the robustness of detection bboxes using the similarity of tracks.
	
	How to filter detection results is tough after the detection process. Many MOT trackers \cite{DBLP:journals/corr/abs-2010-12138,DBLP:journals/corr/abs-2003-13870,DBLP:journals/corr/abs-2006-06664,DBLP:journals/corr/abs-2012-15460,DBLP:journals/corr/abs-2103-08808,DBLP:journals/corr/abs-2004-01888} utilize a simple, threshold-based method, they keep detection bboxes by a threshold i.e. 0.7, then use bboxes as input for similarity calculation, because bboxes with low confidence will contain unrelated information and reduce the score of tracking. Unfortunately, many objects with blur and occlusion may have low confidence values and it is necessary to retrieve them. Thus, to increase the correct detections and make trajectories consecutive we will keep all detection bboxes and process all of them.
	\subsection{Calculating Similarity}
	After the detection part, the detection result will be processed by calculating the similarity between tracklets and detection results, then different strategies will be used to match tracklets and detection bboxes.
	
	Keys for calculating similarities include motion, appearance, and location. Methods like SORT \cite{DBLP:journals/corr/BewleyGORU16} adopt the Kalman filter \cite{kalman1960} for the prediction of tracklets location in the following frames and calculate the IOU distance between detection bboxes and predicted bbox as a similarity value. DeepSORT \cite{DBLP:journals/corr/WojkeBP17} presented extract ReID features of detection results using single ReID model. Some recent methods \cite{DBLP:journals/corr/abs-2012-15460,DBLP:journals/corr/abs-2103-08808,DBLP:journals/corr/abs-2004-01177} utilize designed networks to generate more robust results in situations of large camera motion. Recent researches \cite{DBLP:journals/corr/abs-2010-12138,DBLP:journals/corr/abs-2003-13870,DBLP:journals/corr/abs-2006-06664,DBLP:journals/corr/RenHG015,DBLP:journals/corr/abs-2108-02452,DBLP:journals/corr/abs-2004-01888} utilize joint detection and ReID model because they have higher efficiency. 
	\subsection{Matching}
	After calculating the similarity of bboxes, we will match bboxes and tracklets with strategies.  Traditional methods include Hungarian Algorithm \cite{Kuhn1955Hungarian} or greedy assignment \cite{DBLP:journals/corr/abs-2004-01177}. Some methods like SORT \cite{DBLP:journals/corr/BewleyGORU16} utilize once-matching on detection bboxes and tracklets.  DeepSORT \cite{DBLP:journals/corr/WojkeBP17} utilizes cascaded matching for matching. Some recent methods like \cite{DBLP:journals/corr/abs-2101-02702,DBLP:journals/corr/abs-2105-03247} can utilize attention mechanism \cite{DBLP:journals/corr/VaswaniSPUJGKP17} instead of Hungarian Algorithm for matching process.
	
	In our tracker, we will match our tracks and detections with multiple methods. We will maximize our utilization of bboxes.
	
	\section{SportsTrack}
	
	\subsection{Three-Stage Matching}
	In sports scenes, we will face situations such as motion blur and occlusion in views, and issues such as low confidence caused by motion blur (athletes with motion blur will cause lower confidence value) and incorrect detection results with high confidence (like the single athletes corresponds to many high confidence detection bboxes) will be caused. Some trackers like BOT-SORT \cite{aharon2022bot} utilize a two-stage matching strategy with high and low confidence and they will give low priority to low-confidence detection results, this kind of strategy will affect the matching of more high-quality detection results. Thus, we will present our matching process: Sports Matching. Sports Matching is a three-stage matching process. The first stage is the matching process of all detection bboxes and tracklets with a strict threshold. The second stage is the matching process between high-confidence detection bboxes and tracks. The last stage is the matching process between low-confidence bboxes and tracks. More details will be illustrated in the SportsTrack main process section.

	\subsection{Crowded Track}
	To solve the problem of overlapping targets in the camera view, like the body overlapping caused by the physical confrontation of athletes. Under this situation, detection algorithms can not distinguish multiple athletes and can not detect all athletes in the scene. Based on the investigation of sports matches like basketball, some physical confrontations like boxouts will cause serious long-time body overlapping and will result in the lost of tracking or ID switches. We introduce the term crowded tracks, once track get too close (identified by calculating IOU distance), we will allow one detection bbox to correspond to many crowded tracks. Detailed information is presented as follows, and we will utilize those targets in Sports Matching:
	
	Calculate the IOU between any two unlost tracks, and if the IOU between the
	two targets are greater than 0.45, they are considered crowded tracks, and
	for crowded tracks, find the detection target with the largest IOU, and if its
	IOU is greater than the specified threshold (e.g., 0.6), set the corresponding
	detection bbox as a candidate matching target for the crowded tracks. If tracks are not matched by any detection bbox with the ordinary process, then we will use the candidate detection bbox for matching. 
	
	\subsection{Restroing Edge Tracks}
	
	Some trackers like BOT-SORT \cite{aharon2022bot} and ByteTrack \cite{DBLP:journals/corr/abs-2110-06864} will utilize the Kalman filter to update tracks lost in the frame, and for targets lost at the edge, utilizing Kalman filter will make the predicted position of lost targets far away from our frame. As a result, these targets lost at the edge will never be matched and lost forever. It's reasonable for trackers like BOT-SORT \cite{aharon2022bot} to not process targets lost at the edge since they are usually utilized in the pedestrian scenes, and based on the analysis of pedestrian datasets like MOT20 \cite{DBLP:journals/corr/abs-2003-09003}, its rare for us to find frequent frame-in and frame-out of the same pedestrian. However, the frequent frame-in and frame-out of athletes is common for the sports scene,s and generating incorrect tracklets may affect the result of tracking. Thus during the matching process, we design a method to restore edge-lost tracklets, and the process will be composed of two parts: position calculation and track restoration.
	
	Position calculation: Calculate the position for each lost target and determine whether the position of the lost track is at the center of the image or at the edge of the image, for tracks lost at the center of the image we will use ordinary strategy and for track lost at the edge of the image, we will not utilize Kalman filter, but orientation and ReID feature to restore the track.
	
	Track Restore: If the target is the edge-lost target. If the length of the sequence of unlost tracks is less than 30 and it appears after the lost tracks, and the distance between its appearing angle (taking the image center as the origin) and the lost angle(taking the image center as the origin) of the lost track is less than 90 degrees, it is considered that it may be the same track, and for this case, the ReID distance of two sequences is calculated by taking the lost track the latest 60 ReID records (the actual length may be less than 60), take the latest 10 ReID records of the unlost track, calculate the ReID distance in pairs, count the number of ReID distance less than the specified threshold (e.g. 0.2), if the number is greater than 3, then the two tracking sequences are considered as the same sequence.

	\subsection{SportsTrack Main Process}
	
	The main process of SportsTrack is illustrated as follows.
	
	\begin{enumerate}
		
		\item Calculate the IOU distance between the tracks and the detection target, assume we have 
		M tracks and N detection targets, then the dimensional size of their 
		IOU distance matrix $D$ is $M \times N$.
		
		\item Calculate the ReID distance between the tracks and the detection bboxes, and
		let the feature vector of the ith tracks be $e_i$, the ReID feature vector of the $jth$
		detection target is $f_j$, Then the distance matrix $E$ of ReID is defined as\\
		$$E_{ij}=e_if^T_j$$
		where $T$ denotes the vector transpose operation.
		
		\item Calculate the hybrid distance D1 based on the IOU distance and ReID distance:\\
		$$D1 = \alpha D + (1-\alpha)E$$
		where $\alpha$ is 0.9.
		
		\item Using the Hungarian algorithm matching tracks and detection bboxes with a
		matching threshold of 0.05 and D1 as the loss.
		
		\item For matched tracks-detection targets pairs, update the Kalman filter state of the
		tracks by the corresponding detection bboxes. For unmatched detection targets, they are divided into two groups of high and
		low confidence using a specified threshold (e.g., 0.6) according to their detection confidence value.
		
		\item A new hybrid distance D2 is calculated using the unmatched tracks and the high-confidence detection targets using their IOU distance $D^H$ and ReID distance
		$E^H$:
		$$D2=(1-\alpha)D^H+\alpha E^H$$
		\item Using the Hungarian algorithm matching tracks and detection bboxes with a
		matching threshold of 0.3 and D2 as the loss.
		\item For the tracks-detection target pairs matched in the previous step, update the
		Kalman filter state of the tracks by the corresponding detection target.
		
		\item Further use the unmatched tracking target and the unmatched high confidence
		detection target to calculated a new hybrid distance D3 using its IOU distance $D^{H1}$ and ReID distance $E^{H1}$:
		$$D3=\alpha D^{H1}+(1-\alpha)E^{H1}$$
		
		\item Using the Hungarian algorithm matching tracks and detection targets with a
		matching threshold of 0.7 and D3 as the loss.
		
		\item For the tracks-detection target pair matched in the last step, update the
		Kalman filter state of the tracks by the corresponding detection bboxes.
		
		\item A new hybrid distance D4 is calculated using the unmatched tracking target and
		the low confidence detection target using its IOU distance $D^L$ and ReID distance $E^L$:
		$$D4=\alpha D^L+(1-\alpha)E^L$$
		
		\item Using the Hungarian algorithm matching tracks and detection bboxes with a
		matching threshold of 0.7 and D4 as the loss.
		
		\item For the tracks-detection target pair matched in the previous step, update the
		kalman filter state of the tracks by the corresponding detection target.  For an unmatched tracking target, check its tracking length, and if its starting
		tracking frame number is not the first frame and its length is one, it is considered
		as a mistracked object.
		
		\item For other unmatched tracks, check if satisified: 1.they are `crowded track' 2.candidate matching object bboxes are settled, and if both conditions are satisfied, update
		these unmatched tracking targets with the corresponding candidate matching
		objects and set them to tracking status.
		
		\item The remaining unmatched tracks, if its state in the last frame is the tracking state
		then calculate its missing position, let the image width be $W$, height be $H$, let
		boundary width $b=60$, if the traking object is lost at image’s center area $(b,b,W-b,H-b)$,
		then it is considered lost in the center area of the image, otherwise it is considered
		lost at the edge area of the image, for the tracks lost in edge area calculate its
		missing angle, the calculation formula is:
		$$A=atan2(y-\frac{H}{2},x-\frac{W}{2})$$
		where $x,y$ are the coordinates of the tracks’ centroid.
		
		\item For lost tracks, check the length of time that they have been lost, and if a
		tracking target has been lost for more than 120 consecutive frames, it is
		considered permanently lost and is removed from the tracking list.
		
		\item For the unmatched high-confidence detection bboxes in the previous step, use
		NMS process these detections, the processing threshold is set to 0.45, and next
		the IOU between the remained high-confidence detections and matched
		detection targets in previous steps is greater than 0.45 is removed.
		
		\item The remaining high-confidence detection targets after the previous processing
		step is added to the tracking list as a new tracking target.
		
	\end{enumerate}
	
	\begin{table*}
		
		\caption{Results on SportsMOT validation dataset}
		\label{table:result}
		\centering
		\begin{tabular}{c||c|c|c|c|c|c|c|c|c|c|c|c}
			\hline
			\bfseries Method& \bfseries HOTA & \bfseries DetA & \bfseries AssA & \bfseries DetRe & \bfseries DetPr & \bfseries AssRe & \bfseries AssPr\\
			\hline\hline
			\bfseries SportsTrack (Our Tracker) & 80.863 & 81.294 & 80.466 & 89.593 & 85.982 & 83.448  & 91.847\\
			\bfseries BotSORT \cite{aharon2022bot} & 70.845 & 80.06 & 62.731 & 88.57 & 84.448 & 65.491 & 87.866\\
			\bfseries ByteTrack \cite{DBLP:journals/corr/abs-2110-06864} & 64.571 & 80.396 & 51.928 & 88.647 & 85.873 & 55.897 & 80.966\\
			\bfseries DeepSort \cite{DBLP:journals/corr/WojkeBP17} & 63.957 & 73.823 & 55.479 & 89.942 & 77.426 & 57.508 & 89.713\\
			\bfseries SORT \cite{DBLP:journals/corr/BewleyGORU16} & 58.948 & 71.728 & 48.518 & 89.663 & 75.309 & 51.723 & 81.874\\
			\hline
		\end{tabular}
		
	\end{table*}

	\section{Experiments}
	\subsection{Setting}
	\subsubsection{Datasets}
	We evaluate our SportsTrack on SportsMOT dataset, SportsMOT dataset contains 3 categories, and 240 sports video clips, including sports scenes like basketball, football, and volleyball. The frame rate of videos is 25 FPS, with 720P resolution. Football videos provide outdoor scenes, and volleyball and basketball videos provide outdoor scenes. The average frame of clips is 495 frames. The training dataset includes 45 videos, including 28574 frames. The validation dataset contains 26970 frames. However, the final test set data is inaccessible. To evaluate the performance of SportsTrack compared to other methods, we will utilize the SportsMOT validation dataset for our experiment. 
	
	\subsubsection{Metrics}
	we use HOTA as our evaluation metric. There are many metrics for MOT evaluation like MOTA \cite{journals/ejivp/BernardinS08}, HOTA \cite{DBLP:journals/corr/abs-2009-07736} and IDF1 \cite{DBLP:journals/corr/RistaniSZCT16}, MOTA is more focusing on ID, FP and FN, considering we have a larger number of FP and FN than ID, MOTA will be affected by the performance of detection algorithm. Meanwhile, IDF1 will put more attention to similarity calculation and matching. DetA \cite{DBLP:journals/corr/abs-2009-07736} simply represent the rate of aligning detections and include two sub-metrics: DetRe and DetPr,  and AssA \cite{DBLP:journals/corr/abs-2009-07736} simply calculate the average value of matched trajectory's alignment and include two sub-metrics: AssRe and AssPr. HOTA will have a better tread-off between detection, matching, and localization performance.
	
	\subsubsection{Object detection model}
	We use YOLOX\cite{yolox2021} as our detection model,
	we use the yolox-x configuration, the image input size is $1440 \times 800$, we use the official weight
	trained on COCO as pre-training weight, we only train the head, the backbone and neck are
	frozen during training, the training data are sportsmot train and val dataset, training
	epochs=50, batch size=40, the learning rate strategy is yoloxwarmcos, the initial learning rate
	is $\frac{0.01}{64}$, the optimizer is SGD, momentum=0.9.
	\subsubsection{ReID model for main algorithm}
	We use Fast-ReID\cite{he2020fastreid} as our ReID model, of which we
	use the sbs\_S50 configuration, with an image input size $128 \times 384$, we use the official pre-trained weights which is trained on imagenet,
	training data are sportsmot train and val dataset,
	total training 38 epochs, batch size=160, learning rate strategy is CosineAnnealingLR, the initial
	learning rate is 0.00035, the optimizer is Adam, momentum=0.9.
	
	\subsubsection{Keypoints detection model}
	We use hrnet as our keypoints detecter\cite{sun2019deep}, We use the configuration pose\_hrnet\_w48. We use the official pretained weight which
	was trained on COCO.
	
	\subsubsection{ReID model for post-process}
	The post-process ReID model uses deep-person-ReID\cite{torchreid, zhou2019osnet, zhou2021osnet}, of which we use the osnet\_ain\_1x0
	configuration, with an image input size of $128 \times 256$, we use the official pre-
	trained weights which is trained on imagenet, training data are sportsmot train and val dataset
	and dukemtmcreid, total training 300 epochs, batch size=256, learning rate strategy is
	CosineAnnealingLR, initial learning rate is 0.0003, using random\_flip and random\_erase
	transforms, the optimizer is Adam, momentum=0.9, beta1=0.9, beta2=0.99. fixebase\_epoch is 50,
	and total train 50 epochs, open\_layers=classifier; loss function is softmax with label\_smooth.

	\subsection{Evaluation}
	The main evaluation result is shown in Table\ref{table:result}, we reached the 80.863 HOTA on SportsMOT validation datasets. We also reached highest HOTA score 76.264 on the testset of SportsMOT dataset. We will show our evaluation score on validation dataset as follows.

	\section{Conclusion}
	We introduce SportsTrack, a novel tracker for sports scenes, and we have reached 76.264 HOTA on SportMOT. We hope our state-of-the-art tracker can be used in practical applications.

	%\ifCLASSOPTIONcompsoc
	%  % The Computer Society usually uses the plural form
	%\else
	%  % regular IEEE prefers the singular form
	%\fi

	
	
	%\bibliographystyle{IEEEtran}
	%\addbibresource{report.bib}
	%\printbibliography[title=References]
	%\bibliographystyle{IEEEtran}
	%\bibliography{report.bib}{}

	%\bibliographystyle{IEEEtran}
	%\bibliography{IEEEabrv,report}
	
	%\bibliographystyle{apalike}
	%\bibliography{articles}


\bibliographystyle{IEEEtran}
\bibliography{reference}

%\bibliographystyle{IEEEtran}
%\bibliography{IEEEabrv,report}

%\bibliographystyle{apalike}
%\bibliography{articles}

\end{document}